\documentclass{article}
\pdfoutput=1

\PassOptionsToPackage{numbers,sort&compress}{natbib}
\usepackage[preprint]{neurips_2024}
\usepackage[T1]{fontenc}
\usepackage[utf8]{inputenc}
\usepackage{graphicx}
\usepackage{booktabs}
\usepackage{amsmath,amssymb}
\usepackage{microtype}
\usepackage{xcolor}
\usepackage[colorlinks=true,linkcolor=black,citecolor=black,urlcolor=black]{hyperref}
\usepackage{legiblefailures}

\usepackage{numbers}   % generated by analysis/make_numbers.py
\usepackage{tables}    % generated by analysis/make_tables.py
\author{%
  Manas Venkata Sai Ravulapalli\\
  Efficient Computation Inc.\\
  \texttt{manas@perseus.so}
  \And
  Samrath Chadha\\
  Efficient Computation Inc.\\
  \texttt{samrath@perseus.so}
  \And
  Abhinav Hari\\
  Efficient Computation Inc.\\
  \texttt{abhinav@perseus.so}
}

\date{}

\newcommand{\AuthorsPdf}{Manas Venkata Sai Ravulapalli, Samrath Chadha, Abhinav Hari}
\hypersetup{pdftitle={Legible Failures: Detecting and Repairing In-Context Binding Errors},
            pdfauthor={\AuthorsPdf}}

\title{Legible Failures: Detecting and Repairing In-Context Binding Errors}

\begin{document}
\maketitle

\begin{abstract}
\noindent
A wrong answer does not show whether the model lacked the needed information or held it and failed to use it. On an
entity--obligation binding task, a language model can emit an incorrect prompt-supplied binding
while a linear probe can recover the correct one from its frozen hidden state. We measure how often
this occurs across \NModels{} public checkpoints, each evaluated with three seeds. We fit a probe on a training fold, select its layer on a validation fold, and report results on a disjoint test fold. On the trials each model gets wrong, probe accuracy exceeds the strict present-obligation baseline, $1/K = \Present$,
by \GapMean{} (95\% CI \GapCI, bootstrapped over models). A query-entity counterfactual rules out token presence and recency.
A score built from the sign of probe--output disagreement improves failure detection over the model's own confidence by \CertMean{} AUROC (95\% CI \CertCI). Raw probe confidence gives no measurable improvement over model confidence.
Steering the residual stream toward the probe-decoded binding, with no gold label, raises accuracy on all eight models tested by a mean of \SteerMean{} (95\% CI \SteerCI). Where recent studies report that probe-detected errors are resistant to interventions, we find that in-context binding is a setting in which probes are actionable.
\end{abstract}

\section{Introduction}
\label{sec:intro}

External behaviour cannot distinguish two kinds of failure. A model may answer incorrectly
because it lacks a representation of the queried binding entirely, or it may hold a representation in
its hidden state without using it to determine the output. We call the latter a \emph{legible failure}: the
correct prompt-supplied binding remains linearly recoverable even though the model emits the wrong
token. These two failures call for different responses. The first needs the information supplied. The
second may be detectable and repairable at inference time. This paper measures how often legible
failures occur, whether probe--output disagreement can detect them, and whether a
model's decoded state can be used to change the output.

We examine these questions on a controlled in-context binding task. Transformers form structures
that attach attributes to entities \citep{binding2310}, but retrieval from those structures becomes
unreliable as entity lists grow \citep{mixing2510}. Each trial declares $K$ entity--obligation pairs,
inserts an interference block, and queries one entity. The assignments are supplied by the prompt
and resampled independently on every trial, so entity identity alone carries no information about its
obligation. Though \citet{recall2510} argue that
hidden-state probes track recall of parametric knowledge, that explanation is unavailable for these newly sampled bindings.

For each model, we fit probes on a training fold, choose the read layer on a validation fold, and report
results on a disjoint test fold. We then evaluate probe accuracy on the test trials where the model
answers incorrectly. Across \NModels{} public checkpoints, each evaluated with three seeds,
\GapN{} meet the pre-specified failure-count floor. Across those models, probe accuracy on failures
exceeds the present-set baseline by \GapMean{} (95\% CI \GapCI). A query-entity counterfactual
holds the declarations and their order fixed while changing which entity is queried. The probe follows
the corresponding obligation with a mean margin of \CfMean{} (95\% CI \CfCI), showing the probe
reads a query-conditional signal rather than lexical presence or recency. On average over a model's failures, the requested binding remains linearly recoverable from the hidden state.

We next build a failure-detection score from probe--output agreement and compare it with the model's
own confidence, predictive entropy, and sampled self-consistency. The signed score improves failure
detection over the model's confidence by \CertMean{} AUROC (95\% CI \CertCI). Raw probe
confidence adds no measurable improvement, while predictive entropy
remains competitive.

Finally, we test whether the decoded state can be used to change the output. We apply a steering intervention, without a gold label, to the residual stream from the emitted class toward the probe-decoded class at the validation layer. Across all
\SteerN{} models tested, the intervention raises accuracy by a mean of \SteerMean{} (95\% CI \SteerCI). A matched-norm random direction repairs fewer trials, and a gold-target arm bounds the maximum improvement. Prior work finds that probe-detected errors resist intervention \citep{detcorr2604, actionability2603}, but in-context binding is a setting where they are not.

The scope of this paper is deliberately bounded by the synthetic task, the single-token answer, the need for white-box access to hidden states, and the use of frozen checkpoints. We therefore treat the intervention as a causal test of the decoded state under this protocol. Section~\ref{sec:limits} details these limitations.

\section{Experimental setup}
\label{sec:setup}

\subsection{The binding task}

Each trial samples $K = \Kmain$ entities and $K$ obligations without replacement from disjoint
pools, pairs them, and writes the pairs as \texttt{entity: obligation} declarations. A block of
$D = \Dmain$ token IDs drawn uniformly at random from the vocabulary follows, and then the query
\texttt{The task for $e_j$ is:} for a uniformly chosen $j$. The obligation pool holds 25 action
words and the entity pool 47 nouns, filtered per model to those its tokenizer maps to a single
token. Entities and obligations are resampled independently on every trial, so the identity of the
queried entity carries no information about the obligation bound to it.

A trial counts as correct when the highest logit over the obligation pool at the final query
position is $o_j$. Chance under that rule is one over the pool size, about $0.04$. We report
throughout against a stricter reference point, the probability of naming one of the $K$ obligations
actually present in the context, $1/K = \Present$. Quantities in Section~\ref{sec:legibility} are
scored against that stricter number even though the probe faces the same pool-wide decision the
model does, resulting in conservative margins.

\subsection{Probe and failure-conditioned accuracy}

At every layer $\ell$ we standardise the residual stream at the query's final position and fit a
multinomial logistic-regression probe over the obligation pool. The \NTrials{} trials of each run
are shuffled and split three ways. The probe is fitted on the first third, the layer is chosen by
accuracy on the second third, and every reported quantity is computed on the remaining
\NTest{} trials, which no fitting or selection step has seen. At the chosen layer the probe is
refitted on the first two thirds before test-fold reporting.

Layer selection is the step where leakage could enter. A model has tens of candidate layers and
several hundred labelled trials. Choosing the layer on the test fold would tune the reported number
to the reported fold. The validation-chosen layer lies between \LayerFracLo{} and \LayerFracHi{} of
depth and changes across seeds on \LayerMoved{} of \NModels{} models, as a re-run selection step should.

The quantity of interest is $\paccw$: probe accuracy on the test trials whose model output is wrong.
By construction, a model's accuracy is zero on those trials and probe accuracy above the present-set
baseline measures bindings the model held but did not use.

Because the denominator of $\paccw$ is the model's own failure count, an accurate model leaves few trials
to estimate it. Before running the sweep, we fixed a minimum of \MinWrong{} expected wrong test
trials, corresponding to a binomial standard error of $0.091$ at $p = 0.5$. For models below this
threshold, $\paccw$ is not reported. Detection instead uses AUROC over all \NTest{} test trials and
is reported for every model.

\subsection{Models and statistical analysis}

The sweep covers \NModels{} public checkpoints spanning 410M to 14B parameters: the Pythia ladder,
GPT-Neo, OPT, OLMo-2 and OLMoE, the Qwen2.5 ladder, and DeepSeek-Coder, including one base and
instruction-tuned pair. Appendix~\ref{app:models} lists the checkpoints
and layer counts, and Appendix~\ref{app:hyper} the probe hyperparameters and the compute.
All measurements use frozen checkpoints, and we do not fine-tune any model. Checkpoint revisions
were not pinned; reruns against moved default branches therefore need not be bit-identical.

All intervals reported in this paper are 95\pct{} bootstrap intervals.
The model is the unit of generalisation. Every cross-model estimate is a bootstrap over $20{,}000$
resamples of models, averaging a model's three seeds before resampling, so a model contributes
once however many runs it was measured over. Seeds and trials are nested observations. Per-model
counts of the form ``positive on $k$ of $n$'' are descriptive only.

\begin{figure}[t]
  \centering
  \includegraphics[width=\textwidth]{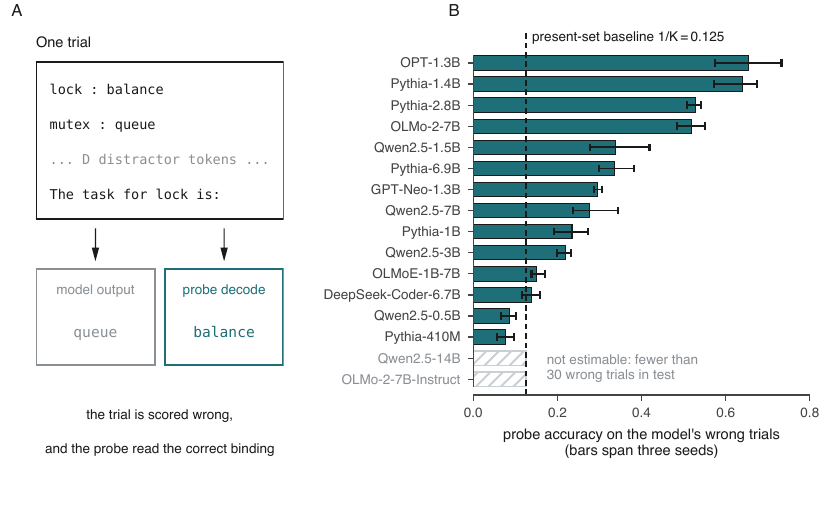}
  \caption{Incorrect outputs retain the correct binding. (A) One trial: $K$ declarations, an interference block,
  and the query. The model emits one obligation and the probe decodes another. (B) Probe accuracy on
  each model's wrong trials, $\paccw$, with the probe fitted on a training fold, its layer chosen on
  a validation fold, and accuracy reported on a disjoint test fold of \NTest{} trials. Bars span
  three seeds. The dashed line is the present-set baseline $1/K = \Present$. Two of \NModels{}
  models fall below the \MinWrong{}-trial failure floor and are hatched; $\paccw$ is not
  reported for them. Across the \GapN{} models for which $\paccw$ is reported, the margin above the
  baseline is \GapMean{} (95\pct{} CI \GapCI, bootstrap clustered on model).}
  \label{fig:legibility}
\end{figure}

\section{Incorrect outputs retain the correct binding}
\label{sec:legibility}

\subsection{Failure-conditioned probe accuracy across sixteen checkpoints}

On the trials where a model emits the wrong obligation, a linear probe recovers the correct one from
the same frozen hidden state (Figure~\ref{fig:legibility}). Across the \GapN{} models for which
$\paccw$ is reported, probe accuracy on wrong trials exceeds the present-set baseline of
$1/K = \Present$ by \GapMean{} (95\% CI \GapCI, bootstrap clustered on model). The median $\paccw$
is \GapMedian{} and the highest is \GapMax{} on \GapMaxModel. On the seed means, \GapPos{} of the
\GapN{} models sit above the baseline. The test folds of these models hold between \WrongLo{} and
\WrongHi{} wrong trials, and every per-model value in Figure~\ref{fig:legibility}B has a denominator
at least twice the pre-registered floor.

The per-model spread is wide, running from a margin near zero on the weakest models to more than
half the range of the metric on \GapMaxModel{} and Pythia-1.4B. A model that fails often also fails
on easy trials. The spread might simply track each model's error rate, but across the \AccGapN{} models for which $\paccw$ is reported, it instead rises with task accuracy
(Spearman $\rho = \AccGapRho$, $p = \AccGapP$). Among the models we can measure,
models that perform better on the task generally have more legible failures, a descriptive correlation.

Choosing the probe layer by test accuracy would fit the reported quantity to the reported fold and inflate
accuracy values. We select the layer based on the validation fold to avoid this. 
On \LayerMoved{} of \NModels{} models, the layer choice moves between seeds.

\CeilHiModel{} and \CeilLoModel{} fall below the failure-count floor. \CeilHiModel{} answers
correctly on \CeilHiAcc{} of trials and leaves about \CeilHiWrong{} wrong trials in a
\NTest-trial test fold. \CeilLoModel{} answers correctly on \CeilLoAcc{} and leaves about
\CeilLoWrong. Their $\paccw$ values are therefore not reported. 
Treating an unreported cell as a failure to exceed the baseline would let two models that contribute no measurement lower the estimate.
The positive correlation above suggests the near-ceiling models, for which $\paccw$ is not reported,
may have the most legible failures.

\subsection{Counterfactual control}

A probe could reach these accuracies by reading which obligations are lexically present, or the most
recent one, without representing any binding at all. The query-entity counterfactual distinguishes
those accounts from a readout of the binding. Holding a declaration fixed, we generate the query for every one of
its $K$ entities in turn and ask whether the probe decode follows the obligation bound to the entity
actually queried. The declared obligations and their order are identical across those $K$
queries, and only the queried entity changes, so a probe reading token presence or recency scores
$1/K$ by construction while a probe reading the binding tracks the query. Across models the
counterfactual accuracy runs from \CfMin{} to \CfMax, exceeding the baseline on \CfPos{} of
\NModels{} models, with a model-clustered margin of \CfMean{} (95\% CI \CfCI). The single model at the baseline, \CfMinModel, is also the one whose
$\paccw$ margin above $1/K$ is most negative (\CfMinGap); the control is null exactly where there is no margin
to explain. Per-model values are in Appendix~\ref{app:results}.

Failed trials therefore retain a query-specific linear signal for the queried binding. The signal
does not determine the output on those trials. The measurement is an average over a model's failures
and identifies no individual trial as legible. Linear
decodability also shows only that the information is present in a form a linear map can read. That
is weaker than evidence that the model uses the information, and
Section~\ref{sec:repair} tests the stronger claim by intervention.

\subsection{Replication on eight further models}

An independent re-implementation, sharing no analysis code with the release pipeline and running
on different hardware, repeated the failure-conditioned probe, counterfactual and detection measurements at $K{=}4$ with no interference block on
eight models spanning four families. The margin above $1/K$ replicates in
all eight: failure-conditioned probe accuracy spans \TrnLegGapLo{} to \TrnLegGapHi{} against the
present-set baseline of $1/K = 0.25$ at $K{=}4$, and the disagreement score separates wrong from correct trials at
AUROC \TrnLegAurocLo{} to \TrnLegAurocHi. The query counterfactual replicates on the
four models it was run on. With the declaration block held bit-identical and only the
queried entity changed, the probe decodes the new entity's obligation at
\TrnQcfFollowLo--\TrnQcfFollowHi{} and the original entity's obligation at only
\TrnQcfStayLo--\TrnQcfStayHi, below the same baseline.

\section{Internal--external disagreement predicts error}
\label{sec:detect}

\subsection{Signed disagreement score}

If the state stays legible on failures, a probe might flag those failures at inference time. The probe's
own confidence is a poor signal. As explained in Section~\ref{sec:legibility}, the probe is often correct
on the trials the model gets wrong. A confident read is not evidence that the output is right. To address this, we score each trial by signed agreement.
For a trial with probe confidence
$\mathrm{conf}_{\text{probe}}$, write
\[
  c \;=\; \mathrm{conf}_{\text{probe}}\cdot\bigl(2\,\mathbf{1}[\text{probe} = \text{output}] - 1\bigr),
\]
such that $c$ is the probe's confidence when probe and output agree and its negation when they
disagree. We evaluate $c$ by the AUROC of separating correct from incorrect trials on the test fold. That
AUROC runs over all \NTest{} test trials of a run, correct and incorrect alike: no comparison in
this section is conditioned on failure, and every model contributes the same number of trials.
We call $c$ a signed disagreement score rather than a certificate as that would imply a formal
property the score does not have.

\subsection{Baselines}

\begin{figure}[t]
  \centering
  \includegraphics[width=\textwidth]{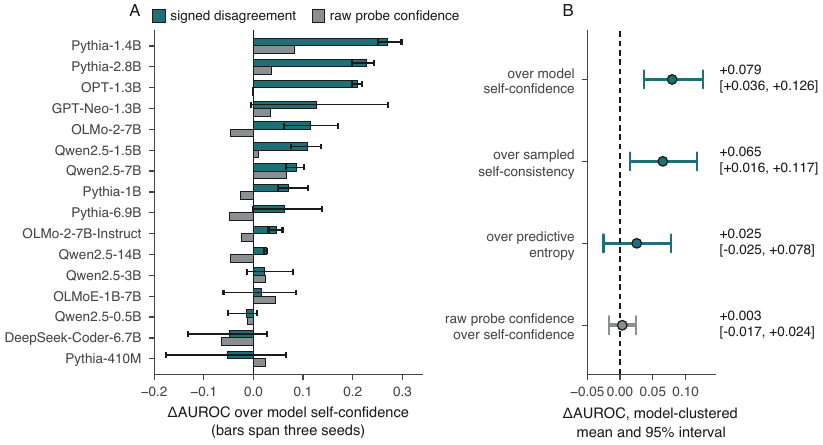}
  \caption{Signed disagreement improves failure detection over self-confidence. (A) Per-model change in AUROC
  relative to self-confidence, for the signed disagreement score and for raw probe
  confidence, on the same test trials with the same probe. Bars span three seeds. (B) Aggregate
  comparisons against four reference detectors, each a mean over \NModels{} models with a 95\pct{}
  interval from a bootstrap clustered on model. The interval against predictive entropy crosses
  zero, and so does the interval for raw probe confidence against self-confidence.}
  \label{fig:detection}
\end{figure}

The reference detector is the model's own confidence, taken as the top-two logit margin over the
obligation pool. Against it, the signed disagreement score improves AUROC by \CertMean{} (95\% CI \CertCI,
clustered on model, and positive on \CertPos{} of \NModels{} models). Raw probe confidence, evaluated
on the same trials with the same probe, improves AUROC by \RawProbeMean{} (95\% CI \RawProbeCI, positive on
\RawProbePos{} of \NModels). That interval crosses zero, thus the probe's confidence alone supplies
no measurable advantage over the model's own confidence. Figure~\ref{fig:detection}A gives the per-model
comparison, and Appendix~\ref{app:results} its values.

Neither of the remaining comparisons needs a probe. Against sampled self-consistency,
the modal answer rate over 16 draws from the model's own obligation distribution, the signed
disagreement score gains \SelfConsMean{} (95\% CI \SelfConsCI). Against predictive entropy over the
same distribution it gains \EntMean{} (95\% CI \EntCI), an interval that crosses zero. Unlike these three
probe-free baselines, which read the output distribution, the score reads the decoded binding.
Section~\ref{sec:repair} uses this quantity as an intervention target.

\subsection{Transfer under distribution shift}

A probe fitted once and then frozen keeps a positive advantage under distribution shift, which is
significant because refitting the detector per deployment condition would be costlier. Across a battery
of six models, the frozen probe's advantage over a model's own confidence is positive on all six under
each of three single-axis shifts: unseen entity and obligation vocabulary, unseen distractor prose,
and unseen interference load. The mean advantage across those three shifts is \TransferMean{} (95\% CI
\TransferCI). When two shifts are stacked, the advantage stays positive on five of six models. \TransferDropModel{}
falls below zero, and it is also the model with the weakest in-domain advantage. Appendix~\ref{app:detect} gives the
per-condition values.

\section{The model's own decoded state repairs behaviour}
\label{sec:repair}

\subsection{Self-gated activation repair}
\label{sec:repair-main}

The probe produces a decode for every trial, and the decoded binding gives an intervention target
that does not require a gold label. At the read layer $\lstar$,
chosen on the validation-fold failures with the test fold untouched by the selection, we take the
class-conditional mean residual
$\mu_c$ of each obligation class over the training fold. On a test trial, write $e$ for the emitted class
and $\hat c$ for the probe decode over the $K$ present obligations. We add
$\alpha\lVert h\rVert\,\mathrm{unit}(\mu_{\hat c} - \mu_e)$ to the residual stream at the query's
final position and re-run the remaining blocks. No gold label enters this arm. The direction is
exactly zero whenever the probe agrees with the output, so the intervention fires only on the
trials the detector of Section~\ref{sec:detect} would flag.

The steering runs use $K = \SteerK$ bindings, a code-like interference block of $D = \SteerD$
tokens, and \SteerTrials{} trials per model, with accuracy scored over the $K$ present obligations
so that chance is $1/K = \SteerPresent$. The task configuration differs from the one in
Section~\ref{sec:legibility}, and accuracies from the two settings are not comparable. We report
$\alpha = \SteerAlpha$. Appendix~\ref{app:repair} gives the sweep over $\alpha$.

\begin{figure}[t]
  \centering
  \includegraphics[width=\textwidth]{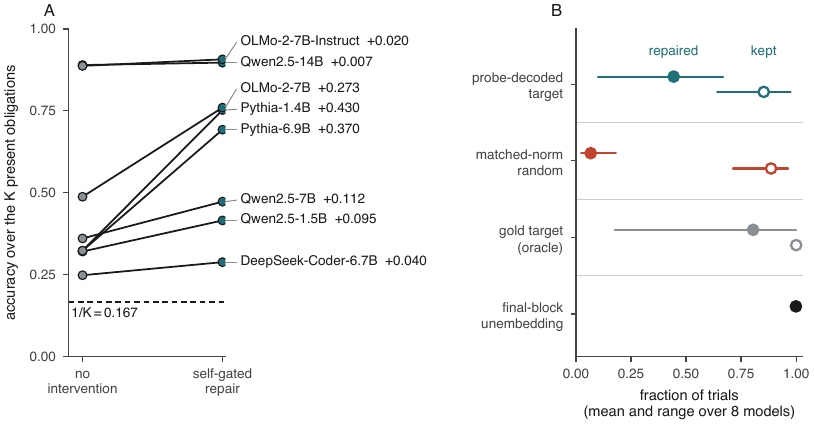}
  \caption{Self-gated activation repair raises accuracy on all eight models. (A) Accuracy without
  intervention and under the self-gated repair, one line per model, at $\alpha = \SteerAlpha$ on
  \SteerTrials{} trials with $K = \SteerK$ and $D = \SteerD$. The decoded arm uses no gold label and
  the direction is zero when probe and output agree. The dashed line is $1/K = \SteerPresent$. (B)
  Control arms over the same eight models. Filled markers, labelled ``repaired'', give the
  fraction of wrong trials an arm repairs; open markers, labelled ``kept'', give the fraction of
  correct trials it preserves. Each is a mean with the range across models. The gold-target arm bounds what the read layer can do when the target is correct; the
  final-block unembedding arm checks the wiring and is measured on wrong trials only.}
  \label{fig:repair}
\end{figure}

Applied to every test trial, the intervention raises accuracy on all eight models
(Figure~\ref{fig:repair}A and Table~\ref{tab:headline}). The mean change is \SteerMean, with an interval of \SteerCI{} from a bootstrap clustered on model. Of the eight models, two sit near
ceiling and fire on \SteerCeilFiredLo{} and \SteerCeilFiredHi{} of trials. Across the
six models whose baseline accuracy is below $0.85$, the mean change is \SteerHeadMean,
reaching \SteerMax{} on \SteerMaxModel. On those models the direction is non-zero on
\SteerHeadFiredLo{} to \SteerHeadFiredHi{} of test trials. The accuracy change occurs only on
trials the detector flags.

On trials the model already answered correctly, the decoded
arm preserves between \KeepLo{} and \KeepHi{} of them across the eight models. Where the probe misreads
a binding that was correct, the intervention moves the output away from the correct answer.

\subsection{Control arms and seeds}

Figure~\ref{fig:repair}B reports the control arms. A matched-norm random
direction, applied at the same layer on the same trials, repairs \RandLo{} to \RandHi{} of failures
while preserving \RandKeepLo{} to \RandKeepHi{} of correct trials. The decoded and random arms differ
only in direction, so the direction drives the effect. Substituting the gold label for the probe
decode repairs \OracleLo{} to \OracleHi{} of failures, with correct-trial preservation of
\OracleKeepLo{} on every model. That arm bounds what the site can do when the target is right, and
it reaches only \OracleWorst{} on \OracleWorstModel, the model the self-gated arm
helps least (\SteerWorst). Adding the unembedding difference at the final block repairs \WireLo{}
to \WireHi{} of failures, a check that the intervention code functions correctly.

Re-running the four models with the highest failure rates at three seeds each, with
both the trials and the split resampled, gives \SteerSeedList. All \SteerSeedPos{} of
\SteerSeedRuns{} seed runs are positive. The least stable is \SeedOracleWorstModel, which also has the
lowest gold-target ceiling of the four (\SeedOracleWorst); the read layer controls less of that model's
decision than it does for others.

A prompt-level arm that re-presents the decoded binding before the query, with no gold label, gives a
convergent result on a separate battery of six models. Appendix~\ref{app:prompt} reports it.

\subsection{Boundary conditions}

The intervention is measured on one synthetic task, at one site, with one direction family. The negative
steering results of \citet{detcorr2604} and \citet{actionability2603} differ on all three aspects, and neither
conclusion transfers to the setting of our protocol. Whether ours would hold under their conditions is
untested, since we did not re-run their tasks. The decode sets what the direction points at, and the gold-target arm shows that the site
itself varies in how much of the decision it controls, from near-total on most models to \OracleWorst{} on \OracleWorstModel. Either factor alone can limit the result,
so we report the two arms separately. We have not resolved the downstream computation that turns a displacement
at $\lstar$ into a change of output.

The re-implementation of Section~\ref{sec:legibility} also bounds the intervention. On the four Pythia
models the decoded-target direction recovers \TrnRepDecLo--\TrnRepDecHi{} of failed trials against a
matched-norm random baseline of \TrnRepRandLo--\TrnRepRandHi, with the no-op arm returning exactly
0.000 recovery and 1.000 preservation on every model. On the four non-Pythia models (0.5B--1.7B) the
same direction recovers at most \TrnNonpyDecHi. The gold-target arm recovers only
\TrnNonpyGoldLo--\TrnNonpyGoldHi{} on those models, so under this protocol the failure sits at
the intervention site and not in the probe: where the gold target cannot repair, no
probe-derived direction can. This does not contradict the eight-model repair result of
Section~\ref{sec:repair-main}, which tunes site and strength per model at larger scales. Rather, it
demonstrates that the intervention does not survive transfer to these models with a fixed site.

The margin above $1/K$ also has a boundary in state type: on mutable variants, where a later instruction overwrites
the binding, and scoped variants, where the binding holds only inside a stated region, failure-conditioned
probe accuracy sits near the present-set baseline at matched difficulty (\TrnMutGap{} and \TrnScopGap{} at
Pythia-1.4B). Stating the overwrite rule in the prompt does not rescue the mutable case (accuracy \TrnExplicitAcc{}
with the rule stated against \TrnImplicitAcc{} without it). The dissociation tracks the state operation,
not task difficulty or an unstated convention.

\begin{table}[t]
  \centering
  \footnotesize
  \caption{The five headline estimates, each with its 95\pct{} bootstrap interval (resampling
  models) and the number of models entering it.}
  \label{tab:headline}
  \setlength{\tabcolsep}{4pt}
  \begin{tabular}{@{}lrrr@{}}
    \toprule
    Quantity & Estimate & 95\pct{} CI & No. models \\
    \midrule
    $\paccw$ above $1/K$        & \GapMean      & \GapCI      & \GapN \\
    Detection over self-conf.   & \CertMean     & \CertCI     & \NModels \\
    Detection over self-cons.   & \SelfConsMean & \SelfConsCI & \NModels \\
    Detection over entropy      & \EntMean      & \EntCI      & \NModels \\
    Activation repair           & \SteerMean    & \SteerCI    & \SteerN \\
    \bottomrule
  \end{tabular}
\end{table}

\section{Related work}
\label{sec:related}

That models represent more than they emit is established for parametric knowledge.
\citet{orgad2410} and \citet{insideout2503} show that internal states carry information about
answers that are not produced by the model. \citet{detext2604} report that answers are recoverable
from reasoning traces before the model emits them. \citet{recall2510} argue that hidden-state probes
largely track recall of parametric knowledge. Our target is an in-context binding whose ground truth
is supplied by the prompt and resampled every trial, which removes parametric recall as an explanation
of the decode. \citet{binding2310} describe the binding-identity structure transformers use to hold
entity--attribute associations. \citet{mixing2510} report that the positional retrieval mechanism
identified for short entity lists becomes unreliable as the list grows, and that models supplement it
by other means.

Probe-based error detection typically reads a property from the internal state alone
\citep{azaria2304,dlk2212}. \citet{dissonance2312} catalogue disagreement between probe output and
model output as a phenomenon.  We use that disagreement as the detector and report that its signed
form outperforms raw probe confidence, which supplies no advantage over the model's own confidence
on our sweep. \citet{sep2406} train hidden-state probes to approximate semantic entropy
from a single generation. Our comparison is against predictive entropy and sampled self-consistency on a
single-token answer, where the advantage over self-consistency excludes zero and the advantage over
entropy does not.

Adding a direction to the residual stream to change a model's output follows inference-time intervention
\citep{iti2306}. \citet{detcorr2604} find that interventions of this family fail to correct hallucinations,
a knowledge--action gap, and \citet{actionability2603} report a comparable failure on a clinical task.
Our setting adds a trial-specific decoded target in place of one global direction, a read layer selected
on validation data along a probe-derived direction, and matched control arms. We differ from those reports on each
of those counts, and record a positive result under our conditions without claiming theirs would behave
the same way. \citet{kappa2509} separately identify knowledge and prediction subspaces in the residual stream
and align them at inference time, a related intervention on a different task.

\section{Discussion and conclusion}
\label{sec:limits}

The task was built for control over the binding, the interference and the baseline. That control is
what lets representation, output and intervention be measured separately on the same trials. The
task is synthetic, the answer is a single token, and the task models maintenance of a binding under
interference. No claim here extends to a naturalistic benchmark. Validating the detector and the
intervention on a real agent trajectory, where a model reads a symbol and then mis-edits it, is
the next test.

Every method here requires white-box access to hidden states at inference time, and both the
detector and the intervention are bounded by probe accuracy. Where the probe cannot read the binding, a
score built from that probe cannot flag failures, and a direction aimed at its decode has nothing
to aim at. Among the four models re-run at three seeds, the one with the lowest gold-target ceiling
is also the least stable.

Because $\paccw$ is not reported for near-ceiling models, the estimate rests on \GapN{} of \NModels{}
models and their rare failures remain untested.
Reaching multi-token settings would need a free-form comparison against semantic entropy.

The intervention is established as a causal effect on this task by its control arms. The account of
why a displacement at $\lstar$ changes the output is incomplete. A direct Jacobian measurement on
the same task finds that the output's sensitivity to the binding direction does not fall on wrong
trials. This rules out one candidate account, that the readout cannot see the binding, without
supplying another.

A subset of failed in-context binding trials retains a query-specific hidden-state signal for the
correct binding, recoverable by a linear probe under a leak-free protocol. Comparing that signal
with the model's output improves failure detection over the model's own confidence, while raw probe
confidence does not, and predictive entropy remains competitive. Adding a residual-stream direction
toward the probe's own decode raises the accuracy point estimate on all eight intervention models,
under controls that separate the direction from a perturbation of the same size. Further testing
would include naturalistic validation on multi-token, free-form agent state.

\section*{Acknowledgements}

We would like to thank Kevin li for his help with proof reading and helping us to better write the paper for readability and suggestions leading to this paper writing. 

\bibliographystyle{plainnat}
\bibliography{references}

\appendix

\section{Models}
\label{app:models}

Table~\ref{tab:models} lists the \NModels{} checkpoints. The layer count is read from the loaded
model configuration at run time and the nominal parameter count is the one in the checkpoint name.
The runs pinned no revision, so each checkpoint is the default branch as of the sweep date. The read layer column gives the layer chosen on the validation fold at each of
the three seeds. It moves between seeds on several models.

\begin{table}[t]
  \centering\footnotesize
  \setlength{\tabcolsep}{2pt}
  \caption{Checkpoints, layer counts, validation-selected read layers, and accuracy on the binding
  task with $K = \Kmain$ and $D = \Dmain$, averaged over three seeds.}
  \label{tab:models}
  \TabModels
\end{table}

\section{Per-model legibility, counterfactual and detection}
\label{app:results}

Table~\ref{tab:results} gives one row per model. Columns two and three are $\paccw$ and the
query-entity counterfactual. For the two models below \MinWrong{} expected wrong trials, $\paccw$
is marked ``not reported''; they contribute to neither the median nor the clustered margin. The counterfactual column
is read against the $1/K$ baseline that a present-token or recency heuristic would score. Columns
four to seven are the detector comparisons behind Figure~\ref{fig:detection}. Model
self-confidence is the top-two logit margin over the obligation pool. Predictive entropy is the
entropy of the softmax over that pool, and sampled self-consistency is the modal answer rate over
16 draws from it. Every column is computed on the same test trials from the same forward pass, so
the comparison holds the trials and the probe fixed.

The number of wrong trials behind each $\paccw$ is \NTest{} times one minus the accuracy in
Table~\ref{tab:models}.

\begin{table}[t]
  \centering\footnotesize
  \setlength{\tabcolsep}{3pt}
  \caption{Per-model legibility, query-entity counterfactual and detector comparisons, over
  three seeds. Columns two and three give the mean over seeds with the range in parentheses.
  Columns four to seven give the change in AUROC of the signed disagreement score against each
  reference detector, except ``raw probe'', which compares raw probe confidence against model
  self-confidence.}
  \label{tab:results}
  \TabResults
\end{table}

\section{Detector transfer under distribution shift}
\label{app:detect}

Table~\ref{tab:transfer} gives the frozen-probe transfer battery. Each column is a distribution
shift applied at evaluation time to a probe fitted in-domain and then frozen: unseen entity and
obligation vocabulary, unseen distractor prose, unseen interference load, and two stacked
combinations.

\begin{table}[t]
  \centering\footnotesize
  \setlength{\tabcolsep}{4pt}
  \caption{Change in AUROC over model self-confidence for a probe fitted in-domain and then frozen,
  under single and stacked distribution shifts.}
  \label{tab:transfer}
  \TabTransfer
\end{table}

\section{Activation repair by model, control arm and seed}
\label{app:repair}

Table~\ref{tab:repair} gives every arm at $\alpha = \SteerAlpha$. ``Rep.'' is the fraction of wrong
test trials the decoded arm repairs and ``Kept'' the fraction of correct test trials it preserves.
``Rand.'' is the matched-norm random direction. ``Gold'' is the same direction family with the
correct label substituted for the decode, and ``Wire'' the unembedding difference added at the
final block.
The steered accuracy column is the self-gated arm applied to every test trial, the number
Figure~\ref{fig:repair}A plots. Table~\ref{tab:alpha} sweeps $\alpha$ above \SteerAlpha. The values
at \SteerAlpha{} are the $\Delta$ and ``Kept'' columns of Table~\ref{tab:repair}.

The seed replication of Section~\ref{sec:repair} reselects the read layer on each seed's validation
failures. The per-seed changes in accuracy are \SteerSeedPerSeed.

\begin{table}[t]
  \centering\small
  \caption{Activation repair at $\alpha = \SteerAlpha$, $K = \SteerK$, $D = \SteerD$,
  \SteerTrials{} trials per model. $\lstar$ is the read layer selected on validation failures.}
  \label{tab:repair}
  \TabRepair
\end{table}

\begin{table}[t]
  \centering\small
  \caption{Intervention strength above $\alpha = \SteerAlpha$. Left: change in accuracy under the
  self-gated decoded arm. Right: fraction of correct trials kept by the decoded arm at the same
  values. Both columns at $\alpha = \SteerAlpha$ are in Table~\ref{tab:repair}.}
  \label{tab:alpha}
  \TabAlpha
\end{table}

\section{Prompt repair and its provenance}
\label{app:prompt}

The prompt-level arm decodes the binding from the model's hidden state at the read layer and
re-presents it before the query, with no gold label. On a separate battery of six models at three
seeds, mean recovery is \RepairMean{} (95\% CI \RepairCI, clustered on model), reaching \RepairMax{}
on \RepairMaxModel, and a format-matched random injection lowers accuracy on \RepairCtrlK{} of
\RepairCtrlN{} runs by \RepairCtrlShift{} on average. Table~\ref{tab:prompt} gives the per-model
values over three seeds on the $K = \Kmain$, $D = \Dmain$ task. The random-injection
control matches the format of the re-presented binding and carries a wrong obligation. The two arms differ only in the content
injected, so the content produces the effect. The gated column intervenes only on trials the detector
flags. The one model with a negative recovery is the near-ceiling instruction-tuned checkpoint,
negative at all three seeds.

This battery was re-run at three seeds for this release and each run's summary is committed. The
per-trial records were not retained, so the numbers in Table~\ref{tab:prompt} are reproducible at
run granularity and not at trial granularity. The activation-repair result of
Section~\ref{sec:repair} carries the causal claim. This arm is reported as a convergent
measurement.

\begin{table}[t]
  \centering\small
  \caption{Prompt-level re-presentation over six models and three seeds. Recovery is
  the change in accuracy relative to no intervention.}
  \label{tab:prompt}
  \TabPrompt
\end{table}

\section{Hyperparameters and compute}
\label{app:hyper}

Probes are scikit-learn multinomial logistic regressions with $C = 0.5$, fitted on standardised
residual-stream activations, with a 400-iteration cap during layer selection and 600 at the chosen
layer. The legibility and detection runs use $K = \Kmain$ and $D = \Dmain$ uniformly
sampled distractor tokens, over \NTrials{} trials shuffled and split three ways into folds of
\NTest{}. The activation-repair runs use $K = \SteerK$ and $D = \SteerD$ code-like distractor
tokens, over \SteerTrials{} trials with the same three-way split. Class-conditional means are computed on the training fold
only, and the read layer is chosen by probe accuracy on the validation-fold failures. Bootstraps
draw $20{,}000$ resamples of models with a fixed seed.

All measurements are forward passes on frozen public checkpoints, run on a single node with eight
NVIDIA H100-80GB GPUs. This paper trains nothing.

\end{document}